\documentclass[letterpaper, 10 pt, conference]{ieeeconf}  

\IEEEoverridecommandlockouts                                                                                        
\usepackage[english]{babel}

\usepackage{graphicx}
\usepackage{animate}
\usepackage{epsfig} 				\usepackage{epstopdf}

\usepackage{listings}
\usepackage{color}
\usepackage{nameref}
\usepackage{hyperref}
\usepackage[colorinlistoftodos]{todonotes}
\usepackage{amsmath}	 			\usepackage{amssymb}  				
\usepackage{dsfont}			\usepackage{mathtools}

\usepackage{epigraph}
\usepackage{lscape}
\usepackage[]{nomencl}				\usepackage{algorithm}
\usepackage{algorithmic}
\usepackage{multicol}
\usepackage{multirow}
\usepackage{etoolbox}

\usepackage{caption}
\usepackage{subcaption}

\usepackage{wrapfig}
\usepackage{multimedia}

\usepackage{siunitx}

\usepackage{floatflt}

\usepackage{url}

\usepackage[absolute,overlay]{textpos}

\usepackage{bibentry}

\usepackage{tikz}
\usepgflibrary{arrows}	

\usepackage{float}
\usepackage[utf8]{inputenc}
\usepackage[english]{babel}
\usepackage{graphics}
\usepackage{animate}

\usepackage{listings}

\usepackage[nostamp]{draftwatermark}

\usepackage{extarrows}

\allowdisplaybreaks[1]

\newcommand{\playvideo}[1]{\href{run:#1}{\includegraphics[scale=0.12]{\RCPath_fig_empty}}}

\usepackage{lipsum}
\usepackage{framed}

\newcommand{\email}[1]{\href{mailto:#1}{\nolinkurl{#1}}}

\newcommand{\link}[1]{\colora{\url{#1}}}

\definecolor{matlab1}{rgb}{0,0,1}
\definecolor{matlab2}{rgb}{0,0.5,0}
\definecolor{matlab3}{rgb}{1,0,0}
\definecolor{matlab4}{rgb}{0,0.75,0.75}
\definecolor{matlab5}{rgb}{0.75,0,0.75}
\definecolor{matlab6}{rgb}{0.75,0.75,0}
\definecolor{matlab7}{rgb}{0.25,0.25,0.25}

\definecolor{darkgreen}{rgb}{0,0.5,0}		\definecolor{purple}{rgb}{0.75,0,0.75}
\definecolor{pink}{rgb}{1,0.4,0.6}

\newcommand{\capitalize}[1]{\expandafter\MakeUppercase\expandafter{#1}}

\newcommand{\colora}[1]{{\usebeamercolor[fg]{framesubtitle}#1}}

\makeatletter
\newcommand*{\compress}{\@minipagetrue}
\makeatother

\renewcommand{\vec}[1]{\boldsymbol{#1}}

\newcommand{\q}{\vec{q}}					\ifdef{\dq}{\renewcommand{\dq}{\dot{\q}}}{\newcommand{\dq}{\dot{\q}}}

\usepackage{array}
\makeatletter
\newcommand{\thickhline}{    \noalign {\ifnum 0=`}\fi \hrule height 1pt
    \futurelet \reserved@a \@xhline
}
\newcolumntype{"}{@{\hskip\tabcolsep\vrule width 1pt\hskip\tabcolsep}}
\makeatother

\AtEndDocument{\par\leavevmode}

\newcommand{\citep}[1]{\cite{#1}}

\title{\LARGE \bf
OmniTact: A Multi-Directional High-Resolution Touch Sensor
}

\author{Akhil Padmanabha$^{1}$, Frederik Ebert$^{1}$, Stephen Tian$^{1}$, Roberto Calandra$^{2,*}$, Chelsea Finn$^{3}$, Sergey Levine$^{1}$
\thanks{$^{1}$Department of Electrical Engineering and Computer Sciences, University of California, Berkeley, CA, USA\newline
        {\tt\small \{akhil.padmanabha, febert, stephentian, sergey.levine\}@berkeley.edu}}\thanks{$^{2}$Facebook AI Research, Menlo Park, CA, USA\newline
        {\tt\small rcalandra@fb.com}}\thanks{$^{3}$Stanford, Department of Computer Science, Palo Alto, CA, USA\newline
        {\tt\small cbfinn@cs.stanford.edu}}\thanks{$^{*}$Work done while at UC Berkeley}}

\begin{document}

\maketitle
\begin{abstract}
	Incorporating touch as a sensing modality for robots can enable finer and more robust manipulation skills.
Existing tactile sensors are either flat, have small sensitive fields or only provide low-resolution signals.
In this paper, we introduce OmniTact, a multi-directional high-resolution tactile sensor.
OmniTact is designed to be used as a fingertip for robotic manipulation with robotic hands, and uses multiple micro-cameras to detect multi-directional deformations of a gel-based skin. This provides a rich signal from which a variety of different contact state variables can be inferred using modern image processing and computer vision methods.
We evaluate the capabilities of OmniTact on a challenging robotic control task that requires inserting an electrical connector into an outlet, as well as a state estimation problem that is representative of those typically encountered in dexterous robotic manipulation, where the goal is to infer the angle of contact of a curved finger pressing against an object. 
Both tasks are performed using only touch sensing and deep convolutional neural networks to process images from the sensor's cameras. We compare with a state-of-the-art tactile sensor that is only sensitive on one side, as well as a state-of-the-art multi-directional tactile sensor, and find that OmniTact's combination of high-resolution and multi-directional sensing is crucial for reliably inserting the electrical connector and allows for higher accuracy in the state estimation task. Videos and supplementary material can be found here.\footnote[4]{\url{https://sites.google.com/berkeley.edu/omnitact}}

\end{abstract}

\section{Introduction}

In order to manipulate an object, a robot needs precise information about the contact state between the gripper and the object being manipulated. While 3D sensors and cameras can provide a global view of a scene, tactile sensors can provide arguably the most direct information about the state of a system, as these sensors can perceive precisely those forces that a robot exerts to manipulate objects. However, the design of effective tactile sensors for robotic manipulation has consistently proven challenging. For a tactile sensor to be useful in robotic manipulation, it must be compact enough to fit inside a robot's finger and must provide a sufficiently rich signal to give the robot relevant information about the contact state. For general-purpose robotic manipulation, it is also crucial for the tactile sensor to be sensorized on as much of the finger's curved surface as possible. A robotic finger with multi-directional sensing can make contact with objects at a variety of points and angles, and therefore should be able to localize objects in a broader range of the state space. 

Most current tactile sensors fall into either of two categories: they provide high spatial resolution on a flat surface, as in the case of the GelSight sensors \cite{johnson2009retrographic, gelsight1, dong2017improved}, or they allow sensitivity on strongly curved surfaces, but with much lower spatial resolution. Curved sensor designs based on capacitive \cite{wettels2009multi}, resistive \cite{cheng2009novel}, optical \cite{piacenza2017accurate, ColumbiaSensor}, sensor arrays have limited spatial resolution due to manufacturing constraints.
High resolution tactile sensing is crucial for high-fidelity manipulation, where precise sensing of the contact state is vital to completing tasks.

In this paper, we propose an optical tactile sensor that uses a similar high-resolution optical sensing principle as the GelSight sensor, but with two crucial differences:
1) Our sensor offers a multi-directional field of view, providing sensitivity on a curved surface.
2) In our sensor, the gel is directly cast on top of the cameras. This results in a more compact form factor compared to previous GelSight sensors, as we entirely remove the support plate and thus eliminate empty spaces between the camera and the gel.

We show that such a sensor can be built using multiple micro-cameras oriented in different directions to capture deformations of a thumb-shaped gel-based skin from all sides. 
In principle, similar micro-camera arrangements embedded into a gel-based skin can be designed for virtually any surface, and could potentially cover a complete robot hand.

\begin{figure}[t]
\centering
\includegraphics[width=\linewidth]{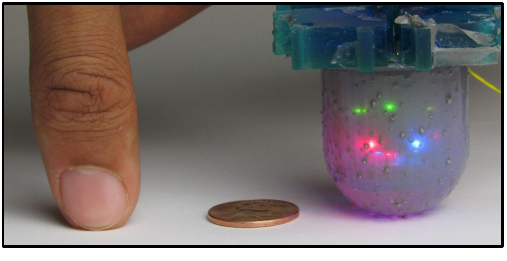}
\caption{Human thumb next to OmniTact, and a US penny for scale. OmniTact is a high-resolution multi-directional tactile sensor designed for robotic manipulation.}
\label{fig:teaser}
\end{figure}

\begin{figure*}[t]
\centering
\includegraphics[width=\linewidth]{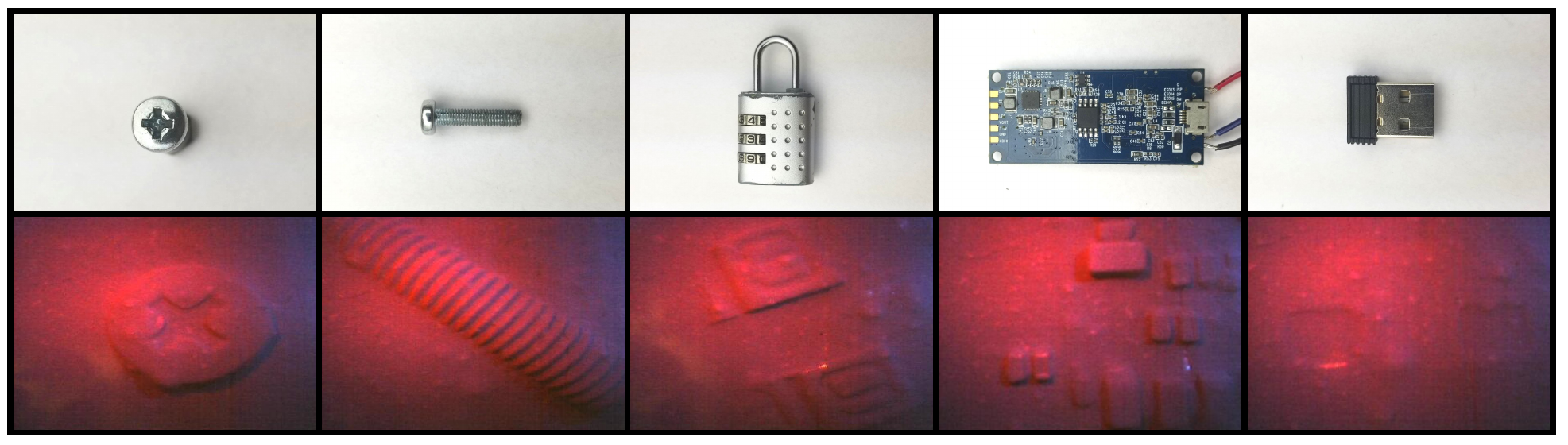}
\caption{Tactile readings from OmniTact with various objects. From left to right: M3 Screw Head, M3 Screw Threads, Combination Lock with numbers 4 3 9, PCB, Wireless Mouse USB. All images are taken from the upward-facing camera.}
\label{fig:example_sensor_readings}
\end{figure*}

\begin{figure*}[t]
\centering
\includegraphics[width=\linewidth]{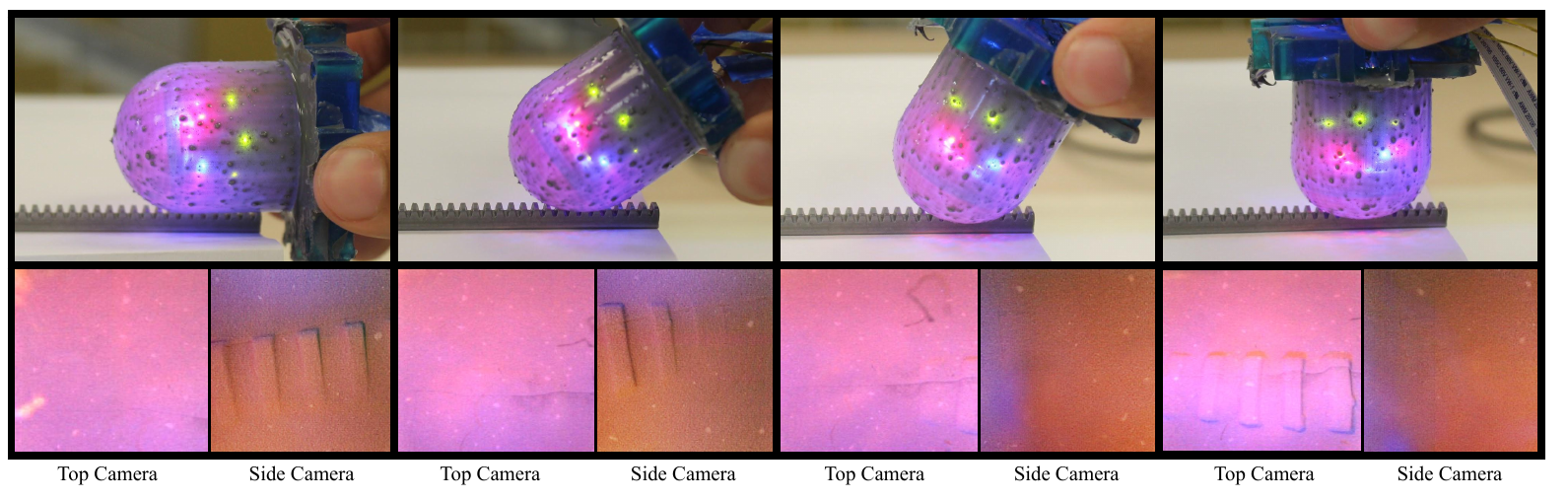}
\caption{Tactile readings from the OmniTact being rolled over a gear rack. The multi-directional capabilities of OmniTact keep the gear rack in view as the sensor is rotated.}
\label{fig:roll_over_gear}
\end{figure*}

Since OmniTact provides an array of high-resolution images (shown in Figures \ref{fig:example_sensor_readings} and \ref{fig:roll_over_gear}), like GelSight sensors \cite{li2014localization, Calandra2017, Calandra2018}, the signals can readily be used in learning-based computer vision pipelines \cite{Tian2019Manipulation}.

This paper presents a novel tactile sensor, shown in \autoref{fig:teaser}, which combines the benefits of multi-directional sensing on strongly curved surfaces with the high resolution and accuracy of optical tactile sensors such as previous GelSight sensors. 
We demonstrate the high spatial resolution and multi-directional sensing capability on a state estimation task, where the sensor is used to estimate the angle of contact when the sensor is pressing against an object.
We also use the OmniTact sensor to solve a robotic control task, where an electrical connector must be inserted into a wall outlet, and observe that OmniTact's multidirectional sensing capability is critical to solving this task.

\section{Related Work}
\label{sec:related}

\begin{figure}[t]
\centering
\includegraphics[width=\linewidth]{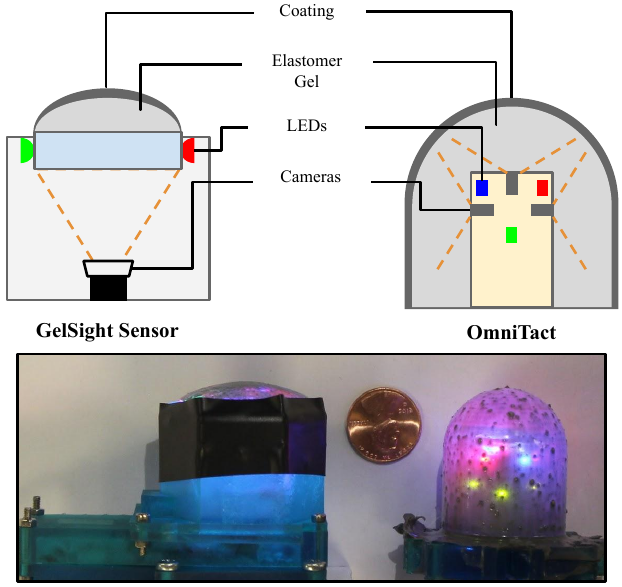}
\caption{Comparison of our OmniTact sensor (right side) to a GelSight-style sensor \cite{dong2017improved} (left side). Using an arrangement of multiple micro-cameras and casting the gel directly on the cameras (without the need for support plates and the empty space between the camera and the gel) allows for a more compact design while enabling a wide field of sensitivity on surfaces with strong curvature.}
\label{fig:gelsight_comparison}
\end{figure}

Our sensor builds on the GelSight design~\cite{johnson2009retrographic} which consists of a camera that captures deformations on an elastomer slab.
As illustrated on the left side of \autoref{fig:gelsight_comparison}, the gel slab of the GelSight sensor is coated with a layer of reflective paint and illuminated by LEDs of different colors. 
A key advantage of GelSight sensors over other tactile sensors is that the images provide a rich signal from which a wide range of relevant information, such as object geometry \cite{li2014localization}, grasp stability \cite{Calandra2017, Calandra2018} and object hardness \cite{yuan2016estimating} can be extracted.
The images from the camera can easily be used with standard convolutional networks \cite{ Calandra2017, Calandra2018,yuan2016estimating}, which have been tremendously successful in computer vision~\cite{krizhevsky2012imagenet}.
Despite these advantages, previous GelSight sensor designs \cite{li2014localization} are bulky (\SI{35}{\milli\meter} x \SI{35}{\milli\meter} x \SI{60}{\milli\meter}), making them difficult to apply in a wide range of applications.
The GelSlim sensor \cite{donlon2018gelslim} integrated mirrors and light guides to make the sensor more compact, decreasing thickness to \SI{20}{\milli\meter}. Like this design, OmniTact provides a more slim design (\SI{30}{\milli\meter} diameter), while providing sensitivity on all sides, whereas GelSlim is only sensitive on one side. A unidirectional version of our sensor would only measure \SI{15}{\milli\meter} in thickness.

Sensors that only provide sensitivity on one side restrict the complexity of the tasks that can be performed. While a unidirectional sensor can be mounted inside a parallel jaw gripper, which is sufficient for grasping, it can be difficult to use for more complex tasks that require both localizing objects in the world and perceiving the object that is grasped. A fingertip that is sensitive on all sides can be used on robotic hands in tasks where contacts occur on multiple sides of the fingertip, as illustrated in our experiments.

As discussed by Donlon et al.~\cite{donlon2018gelslim}, integrating multi-directionality into existing GelSight sensor designs is challenging, due to the lack of space on robotic grippers. Our sensor, shown in \autoref{fig:gelsight_comparison} on the right, aims to tackle this challenge using microcameras, allowing for integrated multi-directionality.
Instead of using cameras to sense the deformation of the gel skin, similar multi-directional sensors have been proposed that use arrays of photodiodes \cite{piacenza2017accurate} to support curved surfaces such as robotic fingertips.
However, using single sensing elements, such as photodiodes, does not provide the same spatial resolution as camera chips.
An example of a multi-directional sensor is the BioTac sensor~\cite{Biotac, wettels2011haptic}, which features similar dimensions to a human finger and provides sensitivity on its tip and side. It uses an array of 19 electrodes, a pressure sensor, and a thermistor, providing far lower resolution than cameras used in GelSight sensors or our proposed sensor. 

To evaluate our sensor, we compare it to a conventional flat GelSight sensor and a multi-directional OptoForce sensor on tactile state estimation and control tasks. 
Tactile state estimation has been studied in a number of prior works~\cite{su2015force, wettels2011haptic, eguiluz2016continuous, chen1995edge, suwanratchatamanee2007simple, ito2011contact,reinecke2014experimental}.
In our experiments, we show that our sensor can result in improved performance over a flat GelSight sensor on state estimation, and improves over a curved OptoForce sensor on touch-based control for connector insertion.

\section{Design and Fabrication}

In this section, we provide a summary of the design goals for the multi-directional OmniTact sensor, and then describe the design in detail, including the fabrication methodology.

\subsection{Design Goals}

The main goal of this work is to build a universal tactile sensor that increases the capabilities of robot manipulation by providing more accurate and comprehensive information about the contact state between the robot and its environment. 

\subsubsection{High resolution.} The sensor should provide rich signals from which features relevant for control, such as object positions, can be extracted accurately. Achieving high spatial resolution has proven challenging with capacitive, resistive, or photodiode-based sensors. However, the resolution of camera-based sensors is limited only by the resolution of the camera and the sensitivity of the sensor skin.

\subsubsection{Thumb-like form factor.} It is important for the sensor to fit into robot fingertips. In many cases, the size of the fingertip restricts the possible tasks it can be used for. For example, a large manipulator may have difficulty picking up small or thin objects such as plates or forks.

\subsubsection{Omni-directional sensing.}
Sensitivity on multiple sides enables the estimation of contacts in a wider range of settings. While sensitivity on the inner surface between fingers is necessary for grasping, sensitivity on the other sides can be crucial for localizing objects of interest, or for performing non-prehensile manipulation. 

Motivated by these design decisions, we present the design of the OmniTact sensor in the following section, followed by details of the fabrication process.

\subsection{OmniTact Sensor Design}

Similar to the GelSight sensor \cite{johnson2009retrographic}, our proposed sensor uses cameras to record the deformation of a gel skin coated with a reflective layer (additional details on the GelSight sensor are discussed in Section~\ref{sec:related}).
However, unlike the GelSight sensor, which uses a single camera, our sensor provides for sensing on all sides of a rounded fingertip, using five micro-cameras, as illustrated in Figure \ref{fig:gelsight_comparison} on the right.
Moreover, the gel is not mounted on top of a support plate as in the GelSight, but rather cast directly around the cameras to reduce the size of the sensor.

\begin{table}
\centering
\caption{Key Specifications of manufactured OmniTact prototype. * excluding blind spots.}
\begin{tabular}{l|c}
\hline
Vertical cut field of view* & 270$^{\circ}$ \\
Horizontal cut field of view* & 360$^{\circ}$ \\
Number of cameras & 5\\
Camera resolution & 400x400 pixels \\
Camera frame rate & 30fps \\
Diameter $D$ & \SI{30}{\milli\meter} \\
Height of sensitive area H &  \SI{33}{\milli\meter} \\
\hline
\end{tabular}
\label{tab:specs}
\end{table}

\begin{figure}[t]
\centering
\includegraphics[width=\linewidth]{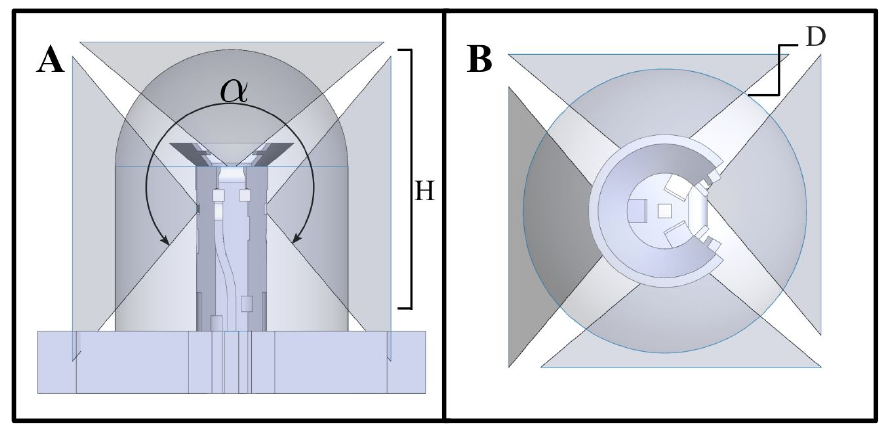}
\caption{Showing the fields of view and arrangement of the 5 micro-cameras inside the sensor. Using this arrangement, most of the fingertip can be sensitized effectively. In the vertical plane, shown in \textbf{A}, we obtain $\alpha=270^{\circ}$ of sensitivity. In the horizontal plane, shown in \textbf{B}, we obtain 360$^{\circ}$ sensitivity, except for small blind spots between the fields of view.}
\label{fig:viewing_angles}
\end{figure}

\begin{figure}[t]
\centering
\includegraphics[width=\linewidth]{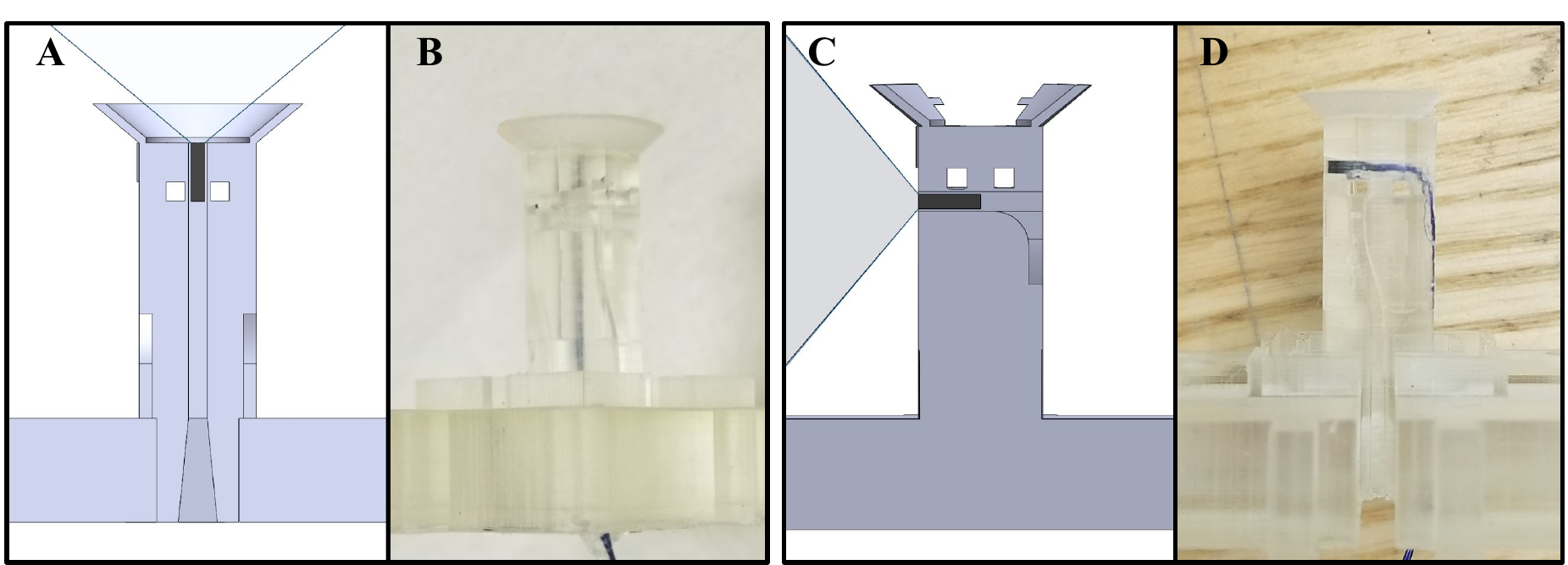}
\caption{The micro-cameras are inserted through channels along the central axis of the camera mount \textbf{A, B} and through the sides of the camera mount \textbf{C, D}.
}
\label{fig:camera_assembly}
\end{figure}

\paragraph{Cameras}
The most important factor determining the size of an optical touch sensor is the choice of camera and the cameras' arrangement relative to the gel. The cameras are chosen to have the maximum possible field of view and the smallest possible minimal focus distance\footnote{The minimal focus distance is the smallest distance between an object and the lens at which the object remains in focus.}.
Cameras with wider fields of view observe larger portions of the sensor's surface area, thus minimizing the total number of cameras required to obtain full coverage of the inner surface of the sensor. Small minimum focus distances reduce the required thickness of the gel skin and overall sensor diameter. 
We found that commercially available endoscope cameras provide the closest match to these requirements and decided to use the 200A CMOS cameras from Shenzhen Eastern International Corporation. Our testing found that each camera has a minimum focus distance of \SI{5}{\milli\meter}, a minimum field of view of about 90$^{\circ}$ and measures \SI{1.35} x \SI{1.35}{\milli \meter} on the sides and \SI{5}{\milli\meter} in length, enabling a very compact arrangement of the cameras. The size of the cameras in relation to the camera mount is illustrated in \autoref{fig:camera_assembly}.
As shown in Figure \ref{fig:viewing_angles}, we arrange the cameras to allow for 270$^{\circ}$ of sensitivity in the vertical plane and 360$^{\circ}$ of sensitivity in the horizontal plane, excluding small blind spots. The blind spots can be reduced in future designs by choosing lenses with slightly larger field of view\footnote{Micro-cameras of similar cost and larger field of view were not available when the prototype was built.}.

\begin{figure}[t]
\centering
\includegraphics[width=\linewidth]{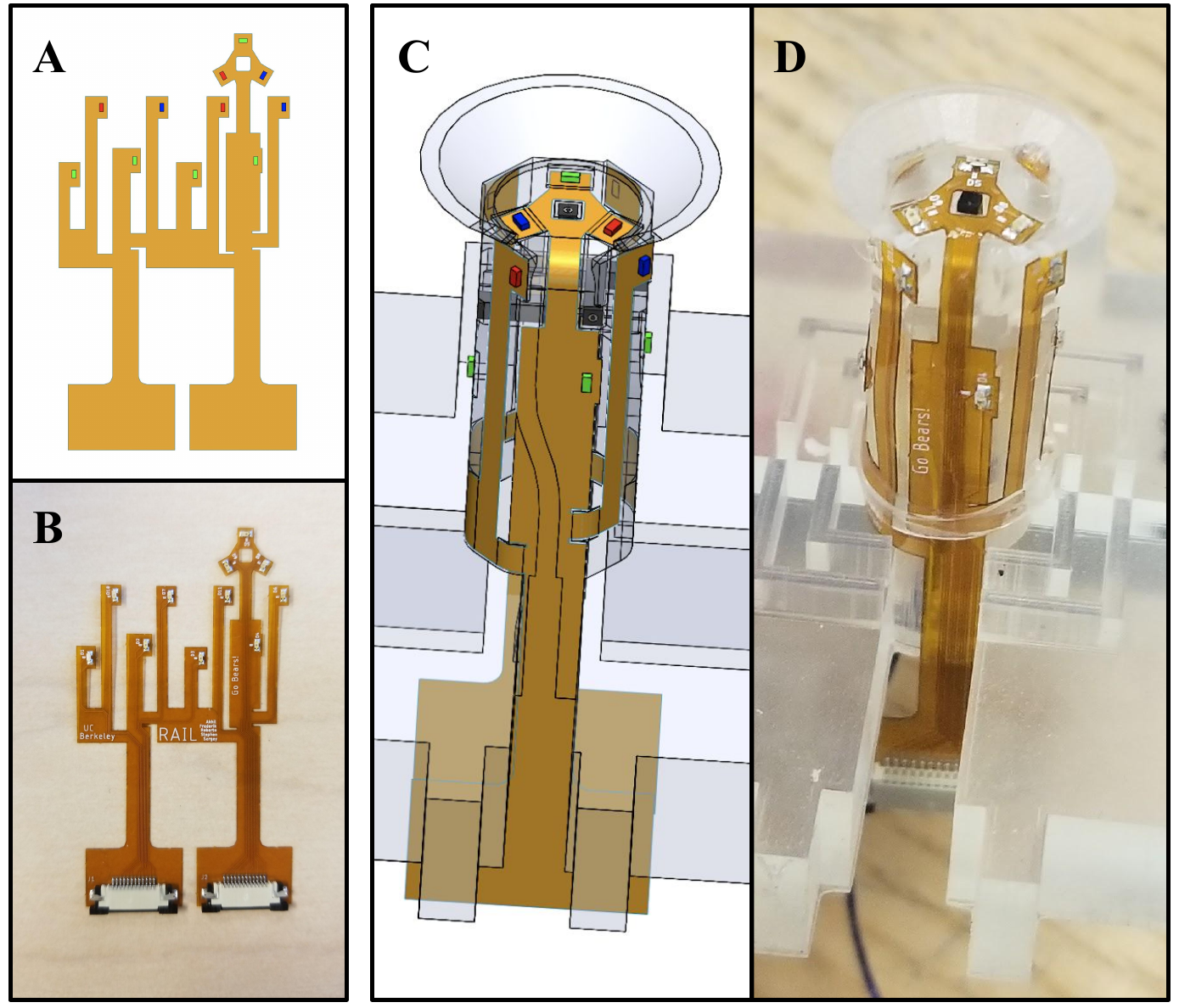}
\caption{\textbf{A} shows the unrolled flex-PCB with the positions of LEDs of different color, \textbf{B} shows the fully assembled flex-PCB. \textbf{C} shows the positions of the LEDs relative to the cameras (in black), and the flex-PCB wrapped around the camera mount. \textbf{D} flex-PCB assembled on camera mount.}
\label{fig:led_placement}
\end{figure}

\paragraph{Illumination}
To optimally detect deformations of the gel in the camera images, the inner surface of the gel should be illuminated as evenly as possible. It is advantageous to illuminate the surface with light of different colors from different directions~\cite{li2014localization}. In principle, this would allow estimating the directions of surface normals \cite{johnson2009retrographic}.
However, in our experimental evaluation detailed in Section~\ref{sec:results}, we show that our sensor can be used for state estimation without explicit estimation of normals or geometry by using a learning-based approach, which directly estimates state from the camera images. As shown in \autoref{fig:led_placement}C, we place 3 colored LEDs adjacent to each camera, which illuminate the gel with red, green, and blue light from different directions. 
The red, blue, and green LEDs are equally spaced from the cameras, ensuring equal distribution of light in the camera image.
In the current design, all LEDs are permanently on, which means that light from different LEDs positioned near different cameras overlaps. 

\paragraph{Camera Mount}
The OmniTact uses a custom-designed camera mount to support the cameras and LEDs for illumination. The mount is designed to minimize blind spots and sensor size, while allowing for easy assembly by hand. The top-facing camera is slid in through the z-axis channel (Fig~\ref{fig:camera_assembly}A), whereas the side cameras are inserted through x and y axis channels (Fig~\ref{fig:camera_assembly}C).
To increase the mechanical stability of the fingertip and the gel, we add a thin cone structure around the top of the camera mount, which also helps reduce interference between the lighting from the LEDs near the top camera and that of the side-facing cameras.

\subsection{Sensor Fabrication}

The camera mount of the sensor, shown in \autoref{fig:camera_assembly}B, D is 3D printed using Formlab's stereo-lithography (SLA) printer, the Form 2, allowing us to print with high resolutions (50 Microns). The Form 2 enables us to manufacture minute features such as the camera channels (\SI{1.45}{\milli\meter} x \SI{1.45}{\milli\meter}). We use a custom designed and manufactured flexible PCB for mounting the LEDs around the cameras.
\paragraph{Assembly process}
The first step in the assembly process is to insert the micro cameras and secure them by gluing the cables down in the channels with E6000 glue, a silicon based adhesive.
The next step is to position, wrap, and glue the flexible PCB around the camera mount. After the glue sets, the camera mount is secured to a mold and filled with silicone rubber. After removing the cured silicone rubber finger from the the mold, the sensor is coated. 

\paragraph{Sensor Coating}
Similarly to \cite{dong2017improved}, we chose \SI{1}{\micro\metre} aluminum powder mixed with the same silicone rubber as used for the gel-skin. A solvent is added to the mix to decrease viscosity. The coating is then completed by pouring the mix over the sensor surface.

\section{Experimental Evaluation}
\label{sec:results}

	The specific tasks that we study include end-to-end control of a robotic arm\footnote{Videos and supplementary material can be found here: \url{https://sites.google.com/berkeley.edu/omnitact}}, for estimating the angle of contact on a flat surface, shown in Figure~\ref{fig:angle_of_attack}, as well as grasping and inserting an electrical connector, shown in \autoref{fig:insertiontask}. For the state estimation task, we compare with a standard GelSight sensor, which is sensorized only on one flat surface. Our specific GelSight sensor is based on the design proposed by Dong et al. \cite{dong2017improved}. Since the OmniTact sensor is nearly symmetric across the four sides, we characterize it using only two cameras -- one of the side cameras and the top camera. For the connector insertion task, we compare OmniTact against a state-of-the-art multi-directional tactile sensor: the single-channel, 3 axis version of the OptoForce.

\noindent \textbf{Neural-network based estimation and control.} 
Both the state estimation and connector insertion tasks use a deep neural network to process the tactile readings, and output a continuous valued output indicating either the robot arm position command or the inferred contact angle. For the state estimation task, our network is based on a modified ResNet-50 architecture \cite{resnet}, where the top layer is removed and the flattened features are fed through 4 fully connected layers with 512, 256, 256, and 256 units, respectively. All other layers in the network are initialized from a model pre-trained on ImageNet, and all layers are fine-tuned during training. For the electrical connector insertion task, the ResNet-18 architecture with the top layer removed is used to generate flattened features, which are fed through 2 fully connected layers of 256 and 64 units, respectively, with ReLU nonlinearities in between. Again, all of the ResNet layers are initialized from a model pre-trained on ImageNet. For the experiments with the OmniTact sensor, the ResNet features produced by the images from all the cameras are flattened and concatenated. The ResNet weights are independently fine-tuned for each camera. The networks are trained with the Adam optimizer~\cite{Kingma2014AdamAM} using a mean-squared-error loss for 100 epochs.

\subsection{Tactile State Estimation - Estimating the Angle of Contact}
\begin{figure}[t]
\centering
\includegraphics[width=\linewidth]{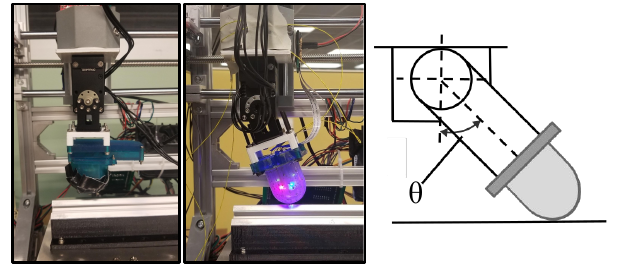}
\caption{Experimental setup for estimating angle of contact~$\theta$ when pressing against a 2020 aluminum extrusion. Left: GelSight sensor. Middle: OmniTact.}
\label{fig:angle_of_attack}
\end{figure}

\begin{table}[t]
\centering
\caption{Angle of contact estimation benchmark for different angle ranges. Numbers are medians with interquartile ranges (IQR) in brackets.}
\begin{tabular}{l |c c c|}
  & \multicolumn{3}{c|}{Median absolute error in $^{\circ}$ (IQR)}\\
  & 0$^{\circ}$  to  22.5 $^{\circ}$  &  22.5$^{\circ}$  to  60$^{\circ}$ & 60$^{\circ}$ to 90$^{\circ}$\\
  \hline
OmniTact (Ours) & 1.142 (1.665) & 1.986 (3.022) & 1.248 (1.683) \\ 
GelSight~\cite{dong2017improved} & 0.325 (0.376)   & 4.228 (6.311) & 1.990 (2.642)\\
\end{tabular}
\label{tab:angle_attack}
\end{table}

In this experiment, we evaluate how well our sensor can estimate the angle of contact with a surface: a basic tactile state estimation task useful in a variety of grasping and manipulation scenarios. To simulate a fingertip contacting a surface at different angles, we set up a state estimation task where we mount the tactile sensor on a rotary actuator attached to the end-effector of a CNC machine. 
The experiment is illustrated in \autoref{fig:angle_of_attack}. 

To collect the data, the tactile sensor is rotated to a random angle in a specified range and the sensor is lowered until it contacts the surface.
Since the travel range of the CNC machine is restricted, we collect data in three different angle ranges, from 0$^{\circ}$ to 22.5$^{\circ}$, 22.5$^{\circ}$ to  60$^{\circ}$, and 60 to 90$^{\circ}$. In each range, we collected 1000 samples, where the rotary actuator is driven to a random angle within the respective range.

The results of the angle estimation task are shown in Table~\ref{tab:angle_attack}. The OmniTact achieves better accuracy than the GelSight sensor in the ranges of 22.5$^{\circ}$ to 60$^{\circ}$ and 60$^{\circ}$ to 90$^{\circ}$. This is expected, since the flat sensorized surface of the GelSight does not cleanly contact the surface at these angles, though the network is still able to perform better than random by picking up on deformations in the plastic sensor housing. These experiments illustrate how a curved finger that is sensorized on multiple sides can enable better state estimation at a wider range of angles.

\subsection{Tactile Control - Electrical Connector Insertion}

\begin{figure}[t]
\centering
\includegraphics[width=\linewidth]{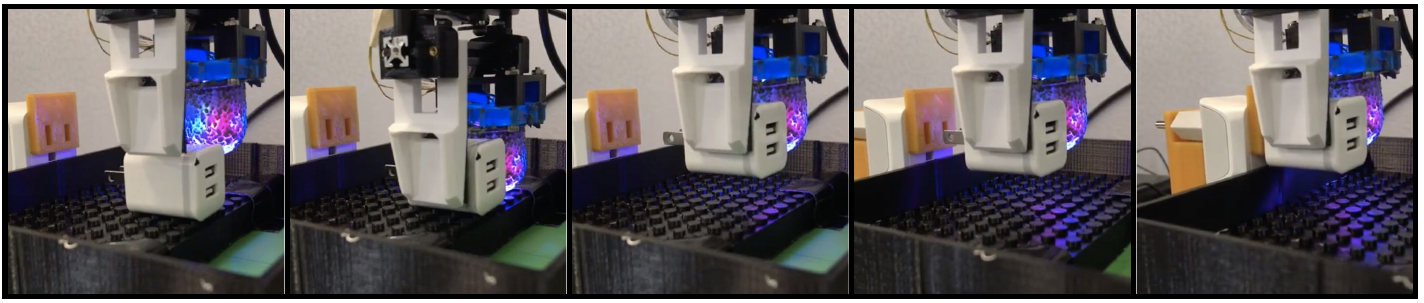}
\caption{Successful insertion of  electrical connector into wall outlet using OmniTact sensor. 
From left to right: 1. Connector is placed between gripper jaws by a human. 2. A random offset is applied to the gripper position. The sensor touches the arena floor and saves a reading from the top camera. 3. Using a pre-scripted pick-up policy, gripper jaws close and connector is lifted. 4. Gripper and connector approach the outlet, and the policy network is queried to determine how to adjust gripper position for insertion. 5. Robot applies adjustment and inserts the connector. 
}
\label{fig:filmstrip}
\end{figure}

In this experiment, we compare how an OmniTact sensor compares with an OptoForce sensor on the task of inserting an electrical connector into a wall outlet solely from tactile feedback (see \autoref{fig:filmstrip} and the appendix for more details). This is a challenging task, since it requires (1) precisely localizing  how the electrical connector is positioned relative to the end-effector, as the way in which the electrical connector is initially placed into the gripper varies, and (2) localizing the wall outlet relative to the robot.

\begin{figure}[t]
\centering
\includegraphics[width=0.95\linewidth]{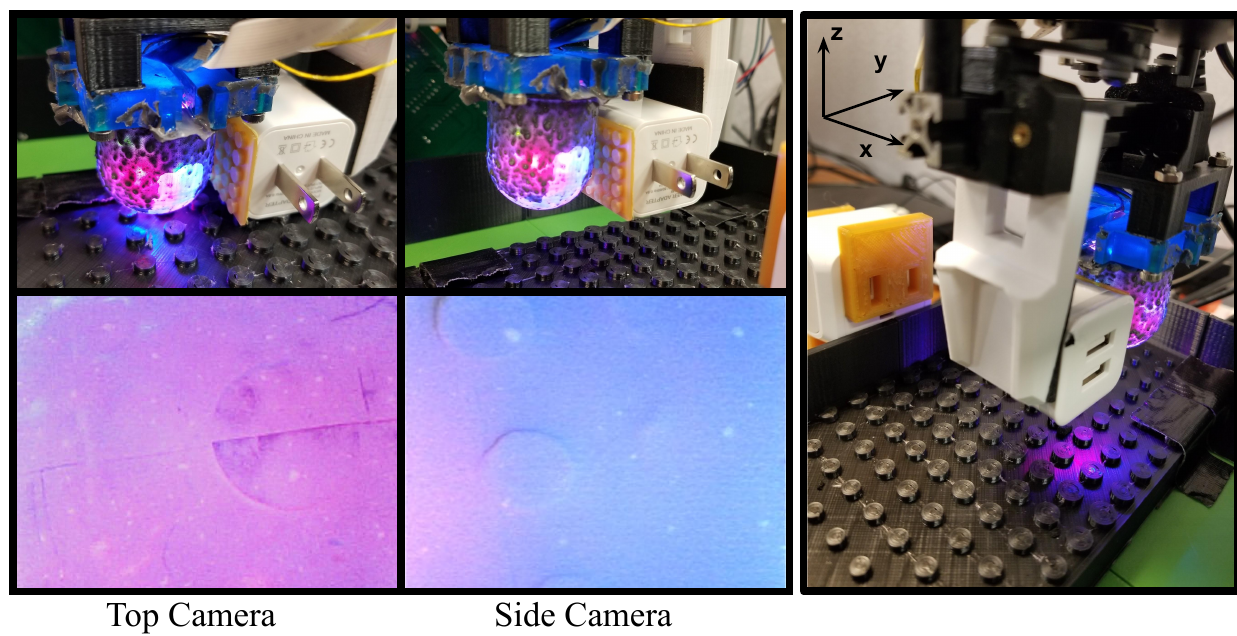}
\caption{Experimental setup for grasping and inserting an electrical connector into a wall outlet. Left: OmniTact touching the bottom plate, causing visible indentation in the top camera image. Middle: Tactile reading from the side-camera after picking up the textured plug. Right: End-effector approaching wall outlet.}
\label{fig:insertiontask}
\end{figure}

In this task, our neural network model directly outputs the desired end-effector position target for a successful insertion. The model is trained on 100 demonstrations of insertions of the electrical connector, provided by commanding the robot's motions via  keyboard control. 
The robot starts off holding the plug a few centimeters away from the outlet, with the tip of the finger contacting a textured floor plate (see Figure~\ref{fig:insertiontask}). In order to correctly determine the insertion pose, the model must determine how it is holding the plug from the sideways-pointing camera, and correctly determine where the end-effector is positioned relative to the outlet by using the pattern on the floor plate to infer the offset.

We compare models that use only the top camera, only the side camera, and both cameras. We also compare to a method that uses inputs from an OptoForce sensor in place of our OmniTact sensor.\footnote{An OptoForce sensor outputs an estimate of the force vector at the tip of its hemispherical sensor pad.} For each combination of sensor inputs, we measure the success rate over 30 trials, only counting full plug insertions as successes.
As listed in \autoref{tab:insertiontask}, using both the tip and side camera of our OmniTact sensor results in the best performance, while using only the top camera results in better performance than using only the side camera. Using only readings from the OptoForce sensor results in the lowest performance (only 5 out of 30 trials are successful).

When using only the top camera, we observe that the robot often fails to fully insert the plug, reflecting the difficulty of estimating the plug's position in the gripper. When only using the side camera, we find that the policy has high error estimating the correct lateral position for insertion. This indicates that the policy uses the top camera to localize the position of the end-effector relative to the plug, and uses the side camera to localize the connector in the gripper, thus utilizing the sensor's multi-directional capability.

\begin{table}[t]
\centering
\caption{Results of electrical connector insertion benchmark (details in Appendix A) showing that including sensor readings from both cameras in the policy inputs outperforms using only a single camera as well as using readings from an OptoForce sensor.}
\begin{tabular}{l | c}
\textbf{Policy Inputs} & Success rate (30 trials)\\
\hline 
OmniTact (Ours) Side Camera only & 50\% \\ 
OmniTact (Ours) Top Camera only & 67\% \\ 
OmniTact (Ours) Side \& Top Camera & \textbf{80}\% \\ 
 \hline
OptoForce & 17\% \\
\end{tabular}
\label{tab:insertiontask}
\end{table}

\section{Discussion and Future Work}
\label{sec:conclusion}

	We presented a design for a multi-directional tactile sensor using multiple micro-cameras to perceive deformations in a gel-coated fingertip. Our design demonstrates that high resolution tactile sensing and the ability to sensorize curved surfaces are \emph{not} conflicting design goals, and that both can be achieved at the same time by using an arrangement of multiple micro-cameras. We further showed how a convolutional neural network can be used to estimate the angle of contact for a finger pressing against a flat surface and perform tactile control to insert an electrical connector into an outlet. Experimental results show that our multi-directional OmniTact sensor obtains high sensitivity for a wider range of angles than a GelSight sensor, and results in higher success rates at inserting an electrical connector purely based on touch sensing. 

A limitation of the current design is the price of the cameras, the most expensive part of the sensor. The endoscope camera used in our sensor cost US\$600 each, for a total cost of US\$3200 for the complete sensor prototype with 5 cameras. This price could be reduced by producing the sensor in larger quantities, or by using different cameras.
However, once sensors of this type can be produced at scale and combined with effective algorithms that utilize touch sensing, future robotic manipulation systems may achieve improved robustness and generality, particularly in delicate and dexterous manipulation settings where direct perception of contacts through touch is critical.

\section*{Acknowledgements} 
We thank Dan Chapman, Kuan-Ju Wu, and Chris Myers from the CITRIS Invention Lab at the University of California, Berkeley for their guidance on designing and manufacturing OmniTact. Additionally, we thank Professor Prabal Dutta, Branden Ghena, and Neal Jackson from Lab11 at the University of California, Berkeley for their electronics advice. This research was supported by Berkeley DeepDrive, Honda, the Office of Naval Research, and the National Science Foundation under IIS-1651843 and IIS-1700697.

\bibliographystyle{IEEEtran}
\bibliography{paper-tactile-servoing}  \clearpage
\section*{Appendix A: Experimental Details for Electrical Connector Insertion Task}
\begin{figure}[h]
\centering
\includegraphics[width=0.55\linewidth]{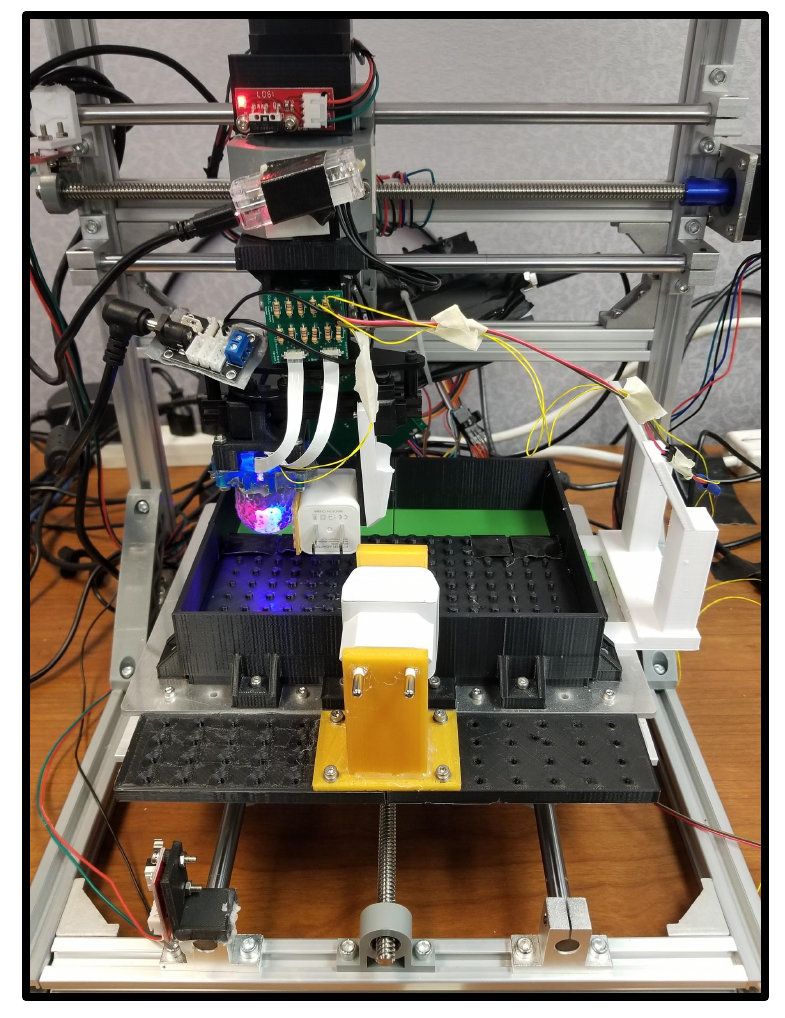}
\caption{Hardware setup for electrical connector insertion task. Modified 3-axis CNC with 3D printed pieces for the arena.}
\label{fig:cnc}
\end{figure}

\begin{figure}[h]
\centering
\includegraphics[width=0.55\linewidth]{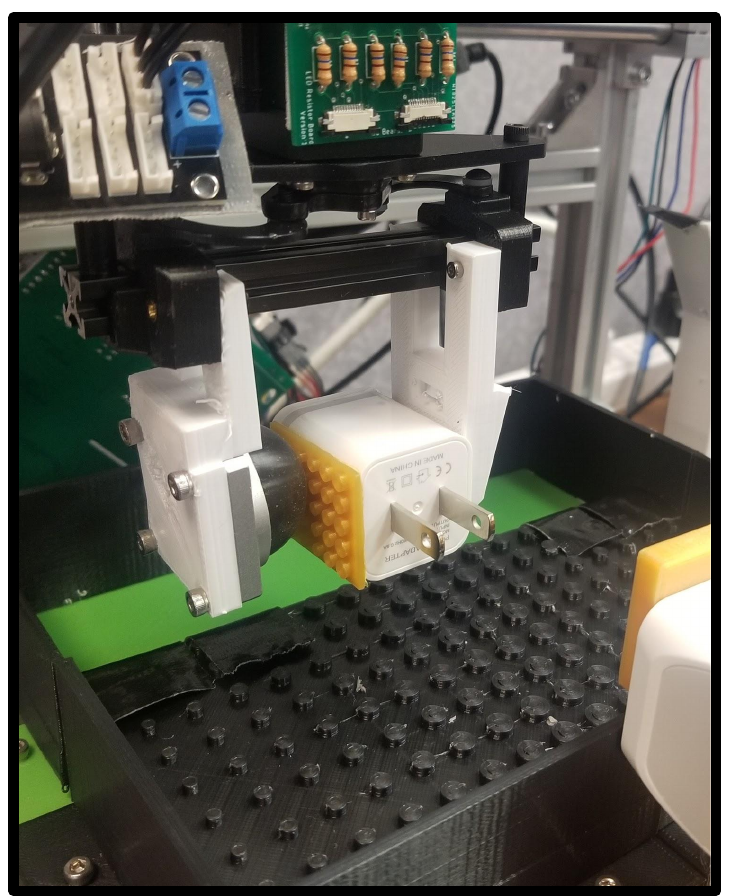}
\caption{OptoForce sensor mounted on the CNC for electrical connector insertion task.}
\label{fig:optoforce}
\end{figure}

\label{appendix:insertiondetails}
For the electrical connector insertion task, we define a successful insertion as one that inserts the prongs of the electrical connector into the outlet, and leaves less than \SI{2}{\milli\meter} between the surface of the outlet and the surface of the connector. For each trial in the experiments, we manually initialize the position of the electrical plug on the textured surface along the axis of insertion (the $x$ direction, as depicted in Figure \ref{fig:insertiontask}) to one of 5 predetermined positions. The set of possible initial positions has a range of \SI{5.5} {\milli \meter}. Additional randomness is added as the gripper position is randomly perturbed before the robot grasps the plug, sampled uniformly in the range of \SI{8} {\milli \meter} in the $x$ direction, and \SI{8} {\milli \meter} in the $y$ direction. We perform 6 trials for each of the 5 predetermined positions, for a total of 30 trials for each experiment.

\end{document}